\documentclass[
]{ceurart}

\usepackage{listings}
\graphicspath{ {./images/} }
\usepackage{siunitx}
\usepackage{nicefrac}
\usepackage{subcaption}
\usepackage{multirow}

\newcolumntype{M}[1]{>{\centering\arraybackslash}m{#1}}
\newcolumntype{R}[1]{>{\raggedleft\arraybackslash}m{#1}}
\newcolumntype{L}[1]{>{\raggedright\arraybackslash}m{#1}}
\def\midtilde@normaltilde{\texttildelow}
\graphicspath{ {./images/} }
\begin{document}

\copyrightyear{2024}
\copyrightclause{Copyright for this paper by its authors. Use permitted under Creative Commons License Attribution 4.0 International (CC BY 4.0).}

\conference{ARISDE2024: 1st International Workshop on Artificial Intelligence for Sustainable Development, 1-3 July 2024}

\title{Computing Within Limits: An Empirical Study of Energy Consumption in ML Training and Inference}

\author[1]{Ioannis Mavromatis}[%
orcid=0000-0002-3309-132X,
email=ioannis.mavromatis@digicatapult.org.uk,
url=https://ioannismavromatis.com/,
]

\author[1]{Kostas Katsaros}[%
orcid=0000-0001-5372-7201,
email=kostas.katsaros@digicatapult.org.uk,
url=https://www.linkedin.com/in/lecost/,
]

\author[2]{Aftab Khan}[%
orcid=0000-0002-3573-6240,
email=aftab.khan@toshiba-bril.com,
url=https://www.linkedin.com/in/aftabkhan00/,
]

\address[1]{Digital Catapult, NW1 2RA, London, U.K.}
\address[2]{Bristol Research \& Innovation Laboratory, Toshiba Europe Ltd., Avon, BS1 4ND, Bristol, UK}

\cortext[1]{Corresponding author: Ioannis Mavromatis, ioannis.mavromatis@digicatapult.org.uk}

\begin{abstract}
Machine learning (ML) has seen tremendous advancements, but its environmental footprint remains a concern. Acknowledging the growing environmental impact of ML this paper investigates Green ML, examining various model architectures and hyperparameters in both training and inference phases to identify energy-efficient practices. Our study leverages software-based power measurements for ease of replication across diverse configurations, models and datasets. In this paper, we examine multiple models and hardware configurations to identify correlations across the various measurements and metrics and key contributors to energy reduction. Our analysis offers practical guidelines for constructing sustainable ML operations, emphasising energy consumption and carbon footprint reductions while maintaining performance. As identified, short-lived profiling can quantify the long-term expected energy consumption. Moreover, model parameters can also be used to accurately estimate the expected total energy without the need for extensive experimentation. 
\end{abstract}

\begin{keywords}
Machine Learning, Power Profiling, Energy Consumption, Sustainable AI, Green AI
\end{keywords}

\maketitle
\section{Introduction}
In recent years, Machine Learning (ML) has seen remarkable advancements, significantly impacting various sectors. However, concerns regarding its environmental footprint have surfaced alongside its growth, showing that 2\% of global carbon emissions will occur from ML pipelines by 2030~\cite{co2Emissions}. The intensive computation of training and deploying ML and Deep Learning (DL) models contribute to substantial energy consumption and, thus, carbon emissions. However, this presents a crucial challenge: how can the ML field continue progressing while aligning with global sustainability goals?

ML is becoming increasingly prominent in current and future use cases and human activities. For example, we see ML playing a pivotal role in 3D-media content supporting Virtual Reality (VR) and Augmented Reality (AR) gaming, interactive art installations, education, etc.~\cite{Moinnereau2022}. We also see AI-native future networks~\cite{aiNative6G} enabling the synergy of AI and data exchange, or we even see the use of Large Language Models (LLMs) and chatbots for everyday activities from millions of people~\cite{genAI}. The flourishing of ML in all domains makes the need for sustainable ML practices not merely a response to environmental concerns but also a strategic imperative for companies and organisations~\cite{mlStrategies}. What is also clear from the above is that ML becomes part of multiple steps in the delivery and operation pipelines of each use case (e.g., the ICT infrastructure, the algorithms, the data processing and analysis, etc.), necessitating ways to reduce energy consumption holistically across the entire system~\cite{sustainable6G}. 




Driven by the above, this paper presents an empirical study of the energy consumption of an immersive media task, i.e., an image classification. Such a task can be common in applications like hand gesture detection, interactive educational games~\cite{imageClassification, educationalGames}, etc. Our investigation aims to offer practical guidelines and best practices that researchers and practitioners can adopt in their applications across the entire ML Operations (MLOps) lifecycle. Even though this is a domain-specific task, our work can be leveraged by ML practitioners aiming for energy-aware optimisations in their ML pipelines across different fields and use cases.

There is recently increased interest in Green and Sustainable ML~\cite{mlStrategies,greenAI}. Sustainable ML practices~\cite{mlStrategies} encompass efficient use of computational resources and holistic optimisation of ML pipelines that collectively lead to reduced energy consumption, minimised carbon footprints, and economic benefits. 
Our paper aims to provide insights into how the ML lifecycle can be optimised for lower energy consumption without compromising performance. We analyse various model architectures and hyperparameters, both for training and inference, to identify areas where energy consumption can be reduced. Based on our findings, we will critically comment on the key contributors to energy reduction and provide ways for estimating the expected energy consumption based on various model parameters. 

The remainder of this paper is structured as follows: Sec.~\ref{sec:related_work} presents recent activities around sustainable ML and discusses their limitations. Green MLOps is described in~\ref{sec:green_mlops}, outlining the energy consumed within an MLOps pipeline. The methodology used for our extensive investigation is illustrated in Sec.~\ref{sec:methodology}. Secs~\ref{sec:results} and~\ref{sec:discussion} present our results and lessons learned. Finally, the paper is concluded in Sec.~\ref{sec:conclusion}

\section{Background and Sustainability Goals}\label{sec:related_work}

The United Nations (UN) has recently presented its 2030 Agenda for Sustainable Development, targeting 17 UN Sustainable Development Goals (SDGs)\footnote{UN Sustainable Development Goals: https://sdgs.un.org/goals}. All 17 SDGs are increasingly important and should be considered by future systems and use cases. Our work is especially linked with the goals below:
\begin{itemize}
    \item \textbf{Goal 9: Industry, Innovation and Infrastructure} - \textit{Build resilient infrastructure, promote inclusive and sustainable industrialisation and foster innovation} - Our work is looking to propose a roadmap for future MLOps frameworks development, paving the way for innovation and good practices across the entire technology stack.
    \item \textbf{Goal 10: Reduced Innequalities} - \textit{Reduce inequality within and among countries} - Reducing energy consumption, ML can become economically viable and sustainable for everyone, fulfilling the 4Cs (Coverage, Capacity, Cost, Consumption) requirements.
    \item \textbf{Goal 12: Responsible Consumption and Production} - \textit{Ensure sustainable consumption and production patterns} - Green ML can significantly reduce the demand for fossil fuel energy production and the total energy consumption.
    \item \textbf{Goal 13: Climate Action} - \textit{Take urgent action to combat climate change and its impacts} - Reducing energy consumption across an entire MLOps pipeline can lead to reduced carbon emissions.
\end{itemize}

Overall, this work aims to provide ways to tackle the energy-hungry tasks of ML training and inference and potential avenues for sustainable MLOps pipelines across the entire computing continuum of any given use case and scenario.

\subsection{Related Work}
Many works present concepts and solutions around Green and Sustainable ML. Some notable examples are~\cite{greenAI,mlStrategies,systematicReview}, where the authors provide statistics on how ML's energy consumption will increase over time. Authors in~\cite{mlStrategies} also compare transformer models running in Google's data centres. All papers discuss the potential benefits of energy reduction from good practices (e.g., early existing, knowledge transfer, etc.) but do not systematically assess those. Our work contributes towards that by conducting an empirical study on real-world hardware.

Researchers have explored various energy reduction algorithms, e.g., pruning~\cite{energyAwarePruning} or quantisation~\cite{energyAwareQuantisation}. These works are smaller-scale investigations and focus on methods that affect the accuracy of a given model. On the contrary, in our large-scale study, we explore ways for energy reduction without changes in accuracy. Similarly to~\cite{erqTradeOffVideoCodecs}, our work will analyse tradeoffs across different configurations and parameters that can be considered when designing ML-enabled systems.

A notable work presenting various measurement campaigns is outlined in~\cite{facebookAI}. The authors focus on how various modifications in an ML pipeline can reduce the environmental impact, targeting system-level holistic optimisations. However, the individual measurements or the models used are not detailed. Our work studies a set of well-known models and datasets to enable readers to understand the differences between distinct hyperparameters and models. Moreover, open-sourcing our code will enable other researchers to replicate our study with different models, datasets or hardware.

The recent literature includes two relevant studies to our evaluation based on real-world measurements~\cite{mlModelSustainability, Strubell_Ganesh_McCallum_2020}. The authors of the first study~\cite{mlModelSustainability} focused primarily on shallow single-layer models. The authors of the second study~\cite{Strubell_Ganesh_McCallum_2020} investigated larger transformer-based models. However, neither includes a deep exploration of how different model characteristics or hyperparameters affect energy consumption. This is a key contribution of our paper.

\section{Green MLOps: A Strategic Imperative}\label{sec:green_mlops}

DevOps merges software development with IT operations to speed up development time using automation and integration tools. MLOps, an extension of DevOps for ML pipelines, focuses on managing the ML model lifecycle efficiently, tackling issues like data management and reproducibility. All ``production systems'' supporting ML-enabled applications usually integrate an MLOps framework~\cite{mlOpsTaxonomy}. Green MLOps extends the idea of MLOps, providing a framework that streamlines ML operations in an energy-aware and cost-effective fashion~\cite{greenAI}.

\subsection{Energy Consumption in MLOps}\label{subsec:energy_consumption}

\begin{figure}[t]
    \centering
    \includegraphics[width=0.7\columnwidth]{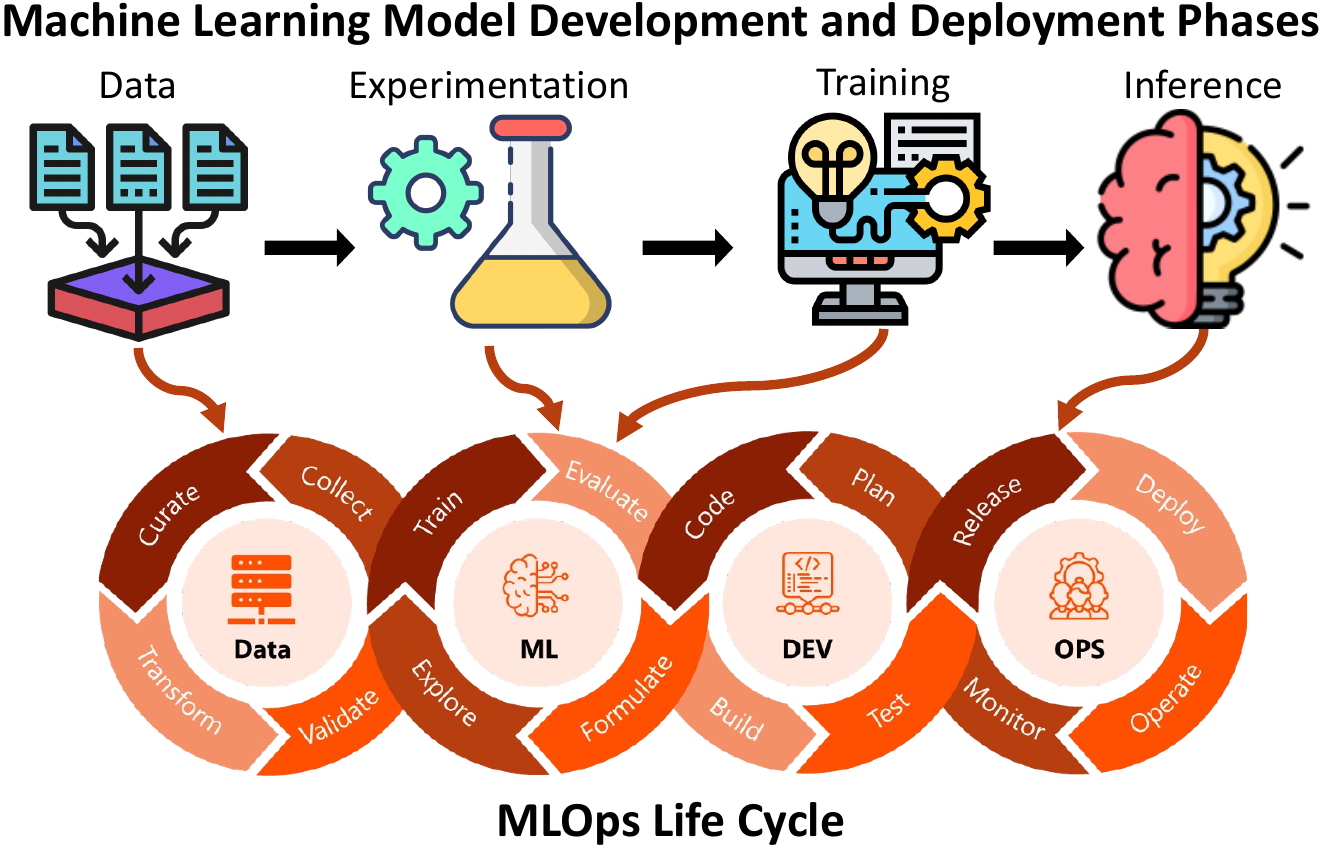}
    \vspace{-3mm}
    \caption{ML model development and deployment phase and the associated MLOps life cycle.}
    \label{fig:high_level}
\end{figure}


MLOps (Fig.~\ref{fig:high_level}) typically involves a \textbf{Data Processing} phase for collecting, curating, and labelling data, and assigning weights to features. In our example image classification task, this might involve creating a dataset of hand gestures. The \textbf{Experimentation} phase follows, where algorithms, model architectures, and training methods are tested. For instance, this could involve testing various models like VGG or YOLO and exploring different hyperparameters, often using grid-search to ensure robustness.


Upon identifying viable solutions, the \textbf{Training} phase involves training the selected models on larger, feature-rich datasets, refining the hyperparameters as needed. After training, models enter the \textbf{Inference} phase. They are deployed to make real-time decisions, e.g., running a hand gesture detection model on a Virtual Reality headset (or a nearby computer). This phase includes continuous performance monitoring and possible re-training.


Training, experimenting, and inferring consume significant energy~\cite{Strubell_Ganesh_McCallum_2020}. Facebook's AI research team~\cite{facebookAI} indicates that inference requires more compute cycles than training, having a split of $10:20:70$ (in \%) between \textbf{Experimentation}, \textbf{Training} and \textbf{Inference}, respectively. Moreover, the energy distribution across the entire MLOps pipeline is roughly $31:29:40$ (in \%) for the \textbf{Data}, \textbf{Experimentation/Training}, and \textbf{Inference} phases. As described in~\cite{Strubell_Ganesh_McCallum_2020}, inadequate hyperparameter tuning and poor neural network management can vastly increase energy consumption by up to $\times2000$ times for Natural Language Model (NLP) models and up to $\times3000$ for a transformer-based NLP.

\subsection{A Comparison between Computation and Data Exchange}

In an MLOps pipeline, energy consumption is the aggregated result of computation and data exchange. Unlike other ICT systems, where computation and data transfer energy use are roughly equal~\cite{energyConsumedNetwork}, ML pipelines are expected to demand more energy for computation. This is because while activities like \textbf{Experimentation} and \textbf{Training} may occur once (e.g., no retraining of the model is required), the \textbf{Inference} and \textbf{Data} phases will always need continuous computation. Additionally, trends like Federated Learning (FL), which distributes training or inference to edge nodes, show even higher energy consumption compared to centralised learning approaches (particularly considering complex ML models) despite the decreased data exchange~\cite{flEnergy}. This is due to the difficulty of parallelising computation, where training and inference are executed across many clients on smaller datasets. 

ML training and inference computation are tightly linked to the model characteristics and hyperparameters, which motivated our investigation. Our paper identifies different model characteristics and hyperparameters that impact the energy consumed within an ML pipeline. We focus our investigation on the \textbf{Experimentation}, \textbf{Training} and \textbf{Inference} phases of an MLOps pipeline and compare parameters such as the model size, the batch size, the time required for training and inference, the Multiply–Accumulate (MAC) operations, the hardware utilisation and the model parameters as a function of the energy consumed. Currently, no study compares the energy consumed by the computation against the data exchange in large-scale ML deployments (e.g., as the one presented from Facebook in~\cite{facebookAI}), highlighting a research gap.

\section{Methodology}\label{sec:methodology}
To investigate the above, measuring the absolute power at frequent intervals and the time required for each experiment is essential. Hardware statistics like the utilisation of resources and the model characteristics should also be captured as part of our experimentation and correlated with the model characteristics and hyperparameters. We implemented a framework that captured all the above and produced the results found in the paper. Our codebase can be found at {\tt\small github.com/ioannismavromatis/sustainable-ai}.

\subsection{Gathering Software-Based Energy Consumption Data}\label{subsec:energyAPIs}
Monitoring energy consumption can be achieved through hardware or software tools. Hardware methods offer precision~\cite{physicalMeter} but face challenges in synchronisation and control~\cite{hardwareSync}, particularly for brief measurements like testing a shallow NN. These methods often require external clocks and costly equipment, limiting accessibility for all ML practitioners. Our investigation employs a software-based approach to measure energy consumption to overcome these issues. This choice not only reduces cost and complexity but also enhances consistency and scalability. Moreover, it allows for parallel evaluations of multiple devices and facilitates testing in complex scenarios, such as FL deployments.

In software-based measurements, power consumption is typically assessed in two ways. The first method estimates power based on a hardware component's Thermal Design Power (TDP) and its utilisation (in a linear relationship). TDP, measured in Watts (\SI{}{\watt}), indicates the maximum power consumption under theoretical full load. However, this method oversimplifies the relationship between power consumption and utilisation~\cite{LIN20211045}, as modern hardware can dynamically adjust the frequency and deactivate entire cores to save energy. A more nuanced approach is based on the hardware's capacitance $C$, voltage $V$, and frequency $f$, as $P = \nicefrac{1}{2} \ C V^2 f$, but obtaining these values for all components is rather challenging. 

As a workaround, manufacturers offer a solution by providing access to energy data through Model Specific Registers (MSRs), like Nvidia's Management Library (NVML) for GPUs and Intel's Running Average Power Limit (RAPL) for CPU and DRAM usage. These methods are reliable with a reported variance of about $\pm \SI{5}{\watt}$ in absolute values while following consistent trends in relative measurements~\cite{Nvidia2016,KatsenouPCS2024}. For consumer CPUs that MSRs do not provide DRAM measurements, DRAM's energy consumption is approximated using $P_{\mathrm{DRAM}} = \sum N_{\mathrm{DIMM}} \times P_{\mathrm{DIMM}}$, where $N_{\mathrm{DIMM}}$ is the number of DIMMs and $P_{\mathrm{DIMM}} = \nicefrac{1}{2} \ C V^2 f$. The operational $V$ and $f$ are accessible from the OS, and $C$ is fixed for all our experiments. This equation is a good approximation as voltage variations during DRAM operations are almost negligible, and operational frequency does not change over time~\cite{dramPowerConsumption}.


\subsection{Calculating Energy Usage in Machine Learning Processes}\label{subsec:energyMeasurements}
As discussed, our investigation will focus on the \textbf{Experimentation}, \textbf{Training} and \textbf{Inference} phases. \textbf{Training} and \textbf{Experimentation} phases are very similar (a model is trained using a set of preconfigured hyperparameters) and thus can be approached similarly in our investigation. To measure the energy consumption we define two metrics, i.e., $E_{\mathrm{tr}}$, which is the total energy consumed during one training session (i.e., for a given model and dataset, with a pre-defined set of hyperparameters and a fixed number of epochs), and $E_{\mathrm{in}}$, which is the total energy during inference (i.e., for a given model and dataset, inferring across all samples with a given batch size). They are as follows:
\begin{equation}\label{eq:training}
    E_\mathrm{tr}=\int^{T_\mathrm{tr}}_{t=0} P_\mathrm{tr}(t) \,dt-\int^{T_\mathrm{idl}}_{t=0} P_\mathrm{idle}(t) \,dt
\end{equation}
\begin{equation}\label{eq:inference}
    E_\mathrm{in}=\int^{T_\mathrm{in}}_{t=0} P_\mathrm{in}(t) \,dt-\int^{T_\mathrm{id}}_{t=0} P_\mathrm{idle}(t) \,dt
\end{equation}
where $T_\mathrm{tr}$ and $T_\mathrm{in}$ are the training and inference times, $T_\mathrm{id}$ is a hardcoded time interval used for the idle experiment, and $P_\mathrm{tr}$, $P_\mathrm{in}$ and $P_{\mathrm{id}}$ are the power measurements during training, testing and when the system is idle. Our framework captures the power consumption at frequent intervals $\Delta t$. Denoting $t_i$ as the $i$-th time interval, the power $P(t_i)$ (this could be either for training or inference) is:
\begin{equation}
P(t_i) = P_\mathrm{CPU}(t_i) + P_\mathrm{GPU}(t_i) + P_\mathrm{DRAM}(t_i)
\end{equation}
where $P_\mathrm{CPU}$, $P_\mathrm{GPU}$ and $P_\mathrm{DRAM}$ are the power consumption, taken in real-time for the CPU, GPU and DRAM, respectively. The energy within $i$-th interval can be calculated as the $E(t_i) = P(t_i)\, \Delta t$. Based on that, the Eqs.~\eqref{eq:training} and~\eqref{eq:inference} can be approximated with the cumulative sum of all intervals, i.e.:
\begin{equation}\label{eq:training_sum}
    E_\mathrm{tr}=\sum^{N_\mathrm{tr}}_{i=0} P_\mathrm{tr}(t_i) \,\Delta t-\sum^{N_\mathrm{id}}_{t=0} P_\mathrm{id}(t_i) \,\Delta t
\end{equation}
\begin{equation}\label{eq:inference_sum}
    E_\mathrm{in}=\sum^{N_\mathrm{in}}_{t=0} P_\mathrm{in}(t_i) \,\Delta t-\sum^{N_\mathrm{id}}_{t=0} P_\mathrm{id}(t_i) \,\Delta t
\end{equation}
where $N_\mathrm{tr}$, $N_\mathrm{in}$  and $N_\mathrm{id}$ are the total number of intervals during training, inference, or idle, respectively. As discussed, data exchange and processing, even though they play a significant role in the energy consumed, will not be considered.

\begin{table}[t]
    \centering
    \caption{Hardware Configurations (HCs). In brackets is the TDP for each hardware component.}
    \begin{tabular}{llll@{}}
    \toprule
    & HC-1 & HC-2 & HC-3\ \\
    \midrule
    CPU$^*$ & i7-8700K (\SI{95}{\watt}) & i9-11900KF (\SI{125}{\watt}) & i5-12500 (\SI{65}{\watt}) \\ \midrule
    \multirow{2}{*}{DRAM} & $4\mathrm{x}\SI{16}{\giga\byte}$ DDR4 & $4\mathrm{x}\SI{32}{\giga\byte}$ DDR4 & $2\mathrm{x}\SI{16}{\giga\byte}$ DDR5 \\
     & \SI{3600}{\mega\hertz} & \SI{3200}{\mega\hertz} & \SI{3200}{\mega\hertz} \\ \midrule
    \multirow{2}{*}{GPU$^+$} & RTX 3080 (\SI{320}{\watt}) & RTX 3090 (\SI{350}{\watt}) & RTX A2000 (\SI{70}{\watt}) \\
     & \SI{10}{\giga\byte} & \SI{24}{\giga\byte} & \SI{12}{\giga\byte} \\
    \bottomrule
    \end{tabular}
    \newline
    $^*$Intel Core, $^+$Nvidia driver v530.30.02, CUDA v12.1
    \label{tab:pcs}
\end{table}

\begin{figure}[t]
    \centering
    \begin{subfigure}{0.99\columnwidth}
        \centering
        \includegraphics[width=\linewidth]{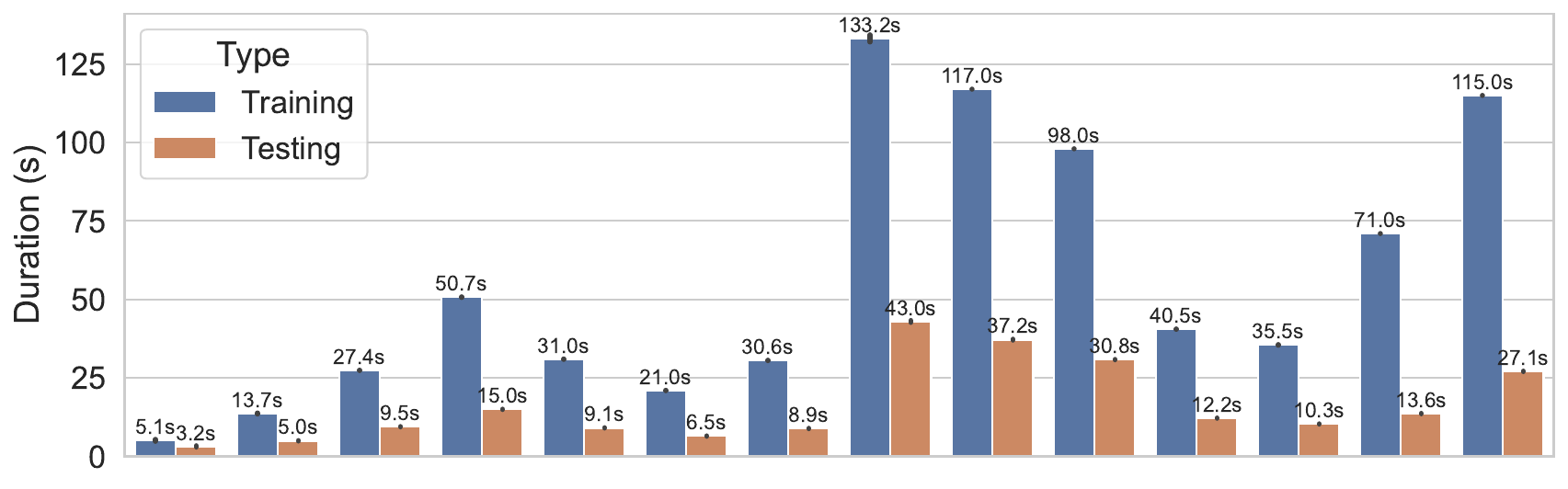}
        \vspace{-6mm}
        \caption{Duration for HC-3.}
        \label{fig:duration_a2000}
    \end{subfigure}%
    \\
    \begin{subfigure}{0.99\columnwidth}
        \centering
        \includegraphics[width=\linewidth]{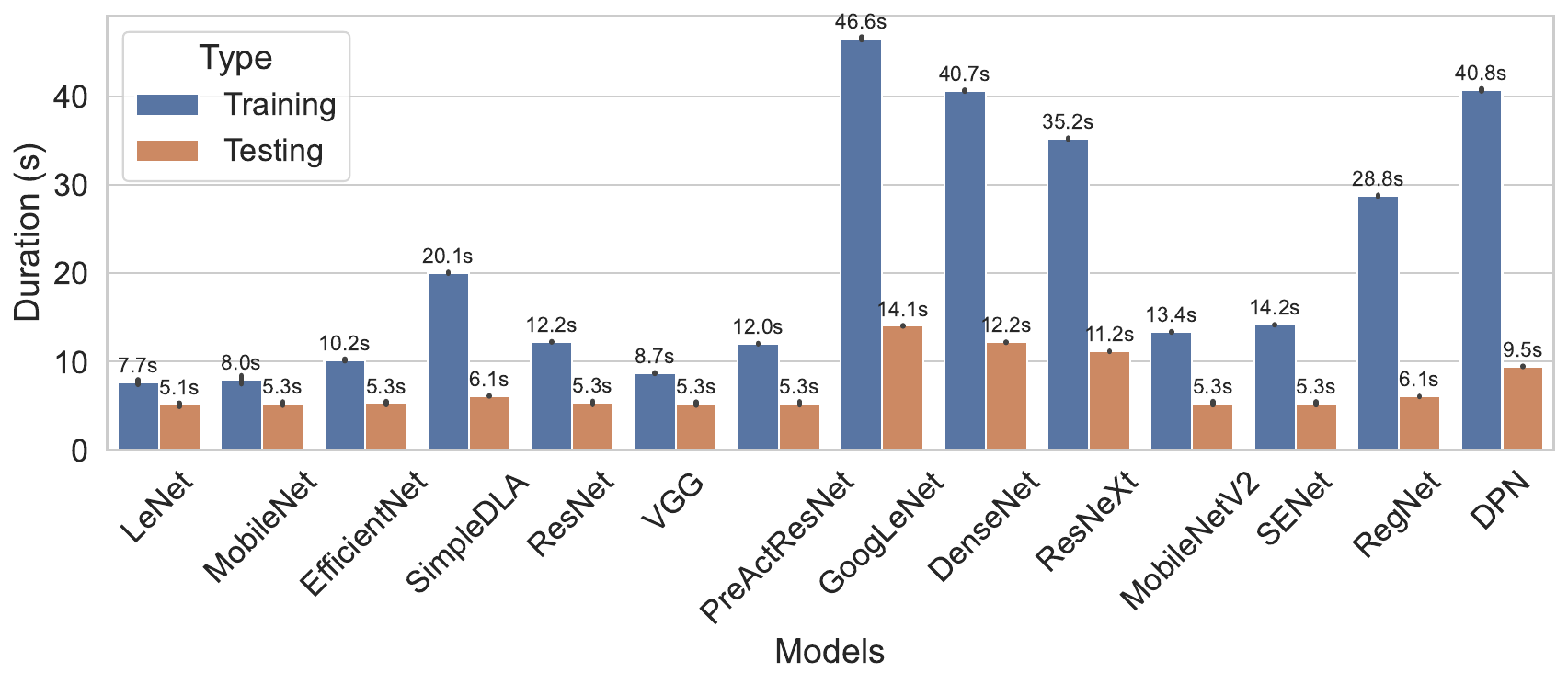}
        \vspace{-6mm}
        \caption{Duration for HC-2.}
        \label{fig:duration_3090}
    \end{subfigure}
    \vspace{-3mm}
    \caption{Training and inference duration (for $50$k samples).}
    \label{fig:combined_duration}
\end{figure}

\subsection{Hardware Stats and Model Characteristics}
Our framework collects utilisation statistics for all resources and model characteristics. The NVML library provides the GPU (and its VRAM) utilisation. For the CPU, the utilisation metrics were directly collected from the OS as a function of each CPU core. The CPU utilisation is calculated as the average utilisation at a given time between all cores. Similarly, DRAM's utilisation was also provided by the OS.

Concerning the model features, several key characteristics emerge as critical indicators when considering the operational efficiency of the model. These include the \textit{model size}, the number of \textit{total and trainable parameters}, \textit{buffer size}, and \textit{MACs}. The model size, measured in bytes (\SI{}{\byte}), is calculated when the model is decompressed and loaded in the VRAM. It includes both the parameters and buffers and represents the overall footprint of the model in memory.

The total number and the trainable parameters are key indicators of a model's complexity. These parameters are different when layers in the model are frozen (i.e., are not updated). A larger number of parameters typically implies a more complex model, which can potentially achieve higher accuracy but at the cost of increased computational resources and memory usage. This complexity can lead to longer training times and may require more powerful hardware.

The buffer size indicates the additional data structures often used for storing intermediate outputs and constants that do not change during training, such as batch normalisation parameters. While they do not directly contribute to the model's learning capacity, they impact the overall memory footprint. A large buffer size can lead to inefficiencies in systems with limited memory.

Finally, the MAC is the fundamental operation in NNs, especially in convolutional layers. The number of MACs provides an estimate of the computational complexity of the model. Higher MACs generally indicate increased computation for both training and inference, leading to longer processing times and increased energy consumption. All the above-mentioned model characteristics are calculated when the model is loaded in the GPU before the execution of each experiment. For our investigation, either independently or as a combination, these parameters will be explored towards total energy consumption.

\section{Results}\label{sec:results}
For our investigation, we consider a simple image classification task. This task was chosen due to the ample models and datasets available in the literature. We conducted a thorough investigation to observe the behaviours of different ML models. In brackets, we present the model variant chosen for our experimentation. We picked: SimpleDLA, DPN (26), DenseNet (121), EfficientNet (B0), GoogLeNet, LeNet, MobileNet, MobileNetV2, PNASNet, PreActResNet (18), RegNet (X\_200MF), ResNet (18),  ResNeXt (29\_2x64d), SENet (18), ShuffleNetV2, and VGG (16), to capture a diverse range of architectures and sizes. All experiments were conducted with the same hyperparameters (batch size of $128$, learning rate $0.001$, SGD optimiser, categorical cross-entropy loss and weight decay $5\times10^{-4}$). We also fixed the seed to ensure consistency across different runs.

We used three different Hardware Configurations (HCs) summarised in Tab.~\ref{tab:pcs}. These three HCs provide diverse playgrounds to explore and identify their differences or similarities and the correlations (Pearson $r$ and Spearman $\rho$) of the different values. As the space in the paper is limited, we will present a subset of the results and discuss the rest in the text. All results can be found in our GitHub repository for further analysis.

Our investigation was based on the CIFAR-10 dataset~\cite{cifar10}. The dataset consists of $60000$ $32\times32$ colour images in 10 classes, with $6000$ images per class. The split between the training and testing set is $50000:10000$. For our evaluation, we replicated the testing set 5 times (i.e., to $50$k samples), so there are consistent samples between training and testing.

\subsection{Initial Statistics}\label{subsec:init_statistics}
Starting with the maximum accuracy reached, most models achieved around $87\%-91\%$ after 100 epochs. The shallower LeNet underperformed as expected, reaching only around $68\%$, whereas MobileNet and EfficientNet reached $81\%$ and $83\%$, respectively. Comparing the time required for training and inference (one epoch of training and $50$k samples of inference), we see the results in Fig.~\ref{fig:combined_duration}. For most models, training takes three times longer than inference due to backpropagation and parameter updating ($r \approx 0.9$ across all models and all HCs). However, as seen, models like DPN, RegNet, etc. do not adhere to this rule of thumb.

\begin{figure}[t]
    \centering
    \begin{subfigure}{0.99\columnwidth}
        \centering
        \includegraphics[width=\linewidth]{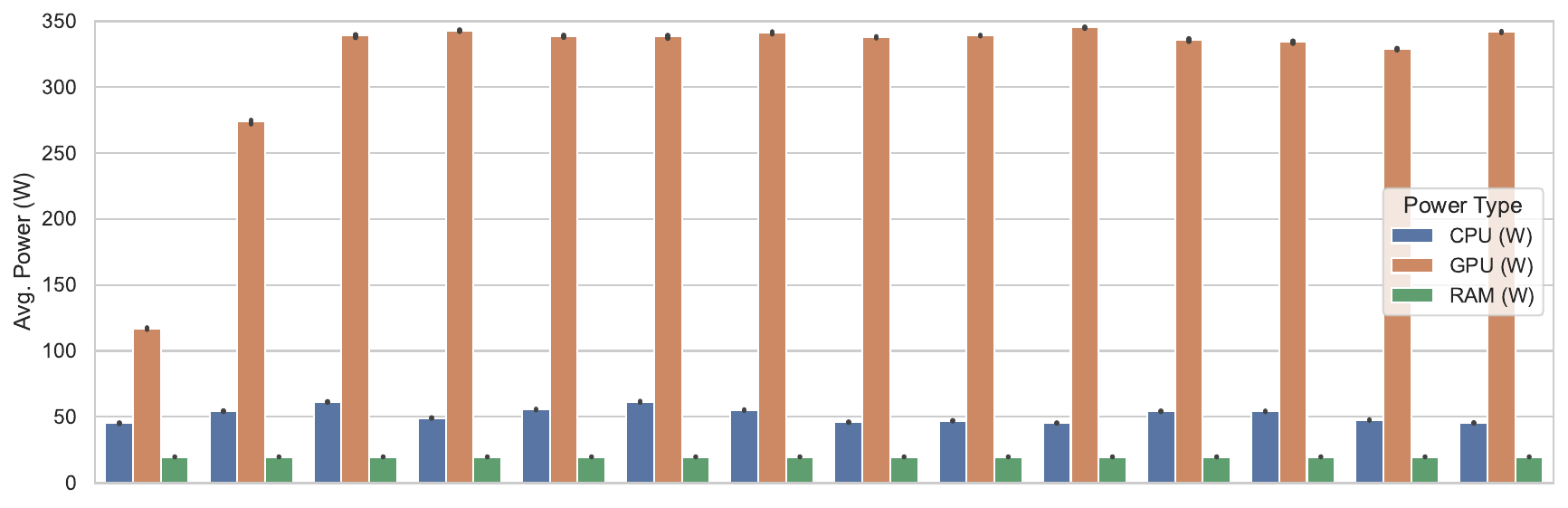}
        \vspace{-6mm}
        \caption{Power usage by model (training).}
        \label{fig:power_training}
    \end{subfigure}%
    \\
    \begin{subfigure}{0.99\columnwidth}
        \centering
        \includegraphics[width=\linewidth]{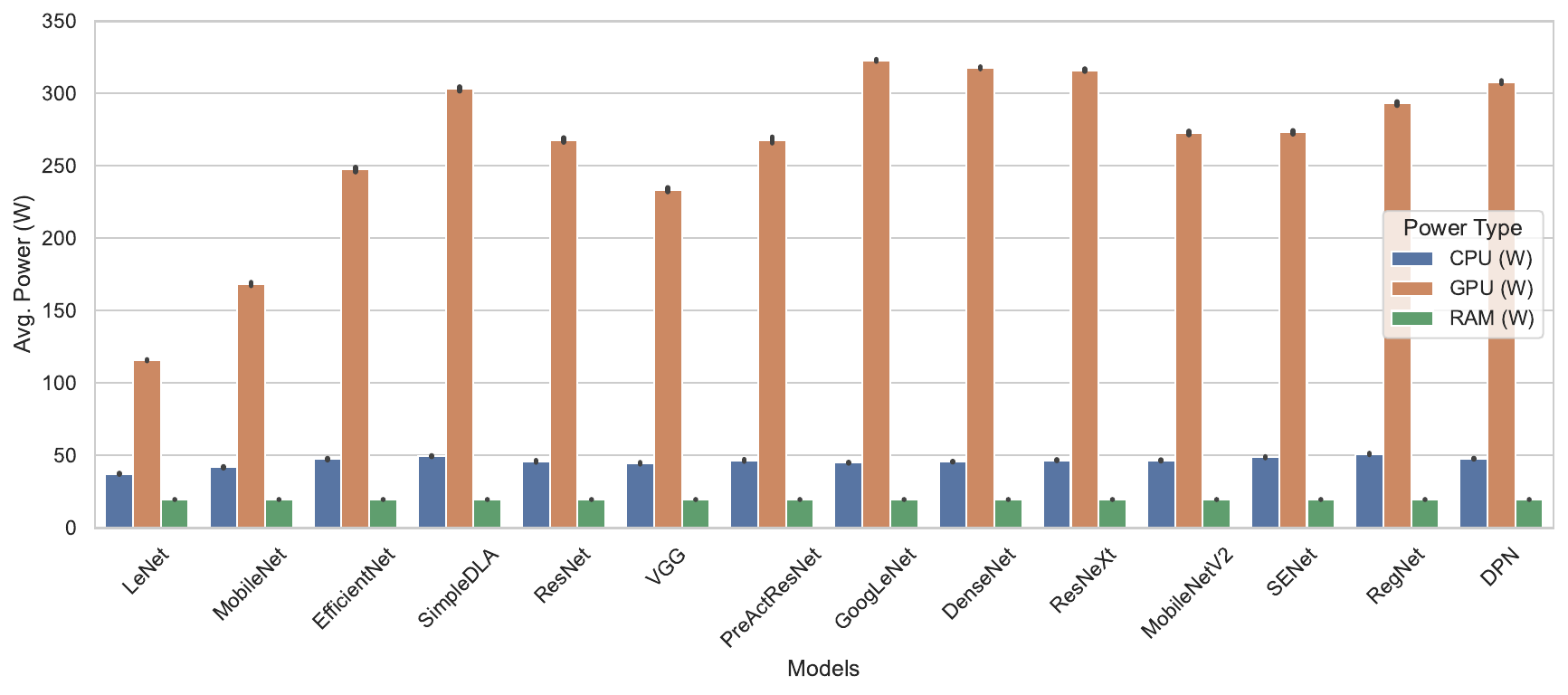}
        \vspace{-6mm}
        \caption{Power usage by model (inference).}
        \label{fig:power_testing}
    \end{subfigure}
    \vspace{-3mm}
    \caption{Average power usage with HC-2.}
    \label{fig:combined_power}
\end{figure}

Across different HCs, we observe large differences in the duration of the same models. For example, PreActResNet at HC-2 (Fig.~\ref{fig:duration_3090}) requires about $5\mathrm{x}$ more time to train or infer compared to LeNet, but at HC-3 (Fig.~\ref{fig:duration_a2000}), that difference goes up to $26\mathrm{x}$. Moreover, even though there is no observable difference during the training phase in terms of time required (relatively -- between models), for inference, we observe that a more powerful GPU (HC-2), when fed relatively small models, parses the same number of samples in about the same time, regardless of the model size. With the inference dictating the energy consumption (as discussed in Sec.~\ref{subsec:energy_consumption}), as a rule of thumb, models achieving similar accuracy but inferring quicker can introduce significant energy benefits in the long term, even if they require more time to train. For example, VGG and ResNet perform equally as well as DenseNet or DPN, but only with a fraction of the energy required, making them better candidates for long-term usage.

\subsection{Power Consumption Measurements}
Fig.~\ref{fig:combined_power} shows the average power consumed for HC-2 for training and inference. All the larger models force the GPU to operate close to its TDP (Fig.~\ref{fig:power_training}). As expected, CPU and DRAM, being underutilised, show roughly equal and not significantly high average power consumption across all models. However, this is not the case for the inference (Fig.~\ref{fig:power_testing}). As shown, many models operate $\geq 30\%$ lower than the GPU's TDP (e.g., VGG), with the CPU and DRAM showing similar results with the training. This is the case for the other two HCs, with the difference being more prominent for HC-1 and less prominent for HC-3.

\begin{figure}[t]
    \centering
    \begin{subfigure}{0.99\columnwidth}
        \centering
        \includegraphics[width=\linewidth]{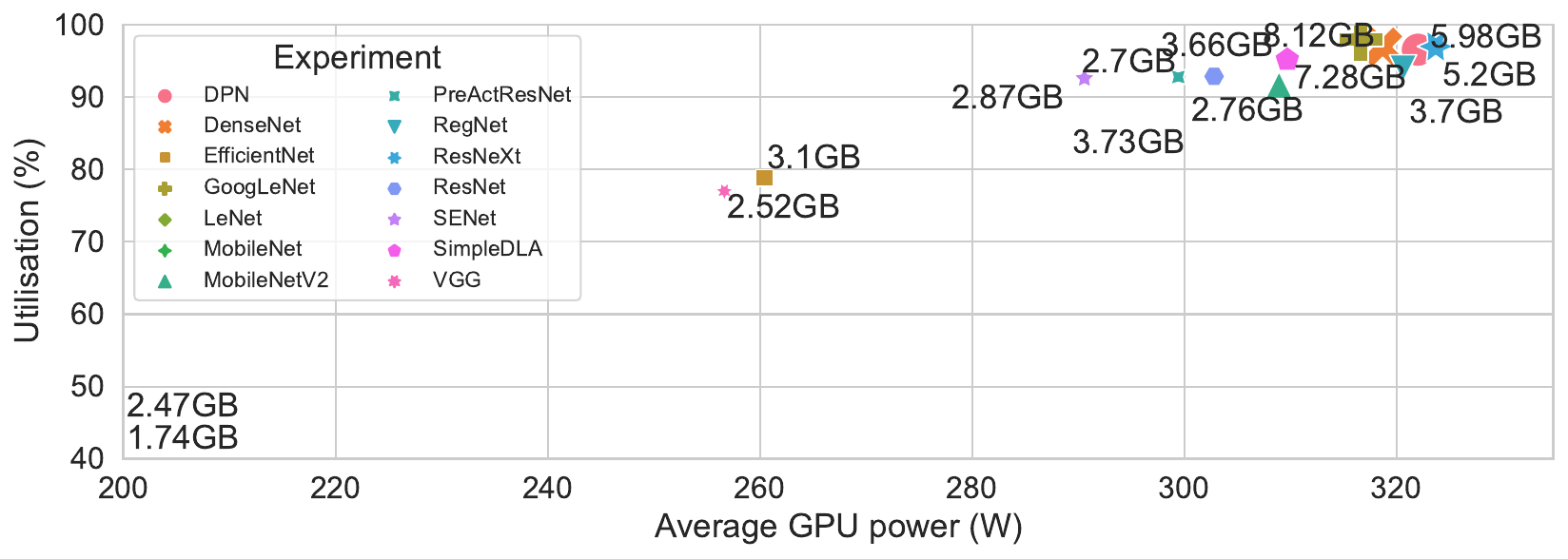}
        \vspace{-6mm}
        \caption{Power usage by model (training).}
        \label{fig:gpu_ram_training}
    \end{subfigure}%
    \\
    \begin{subfigure}{0.99\columnwidth}
        \centering
        \includegraphics[width=\linewidth]{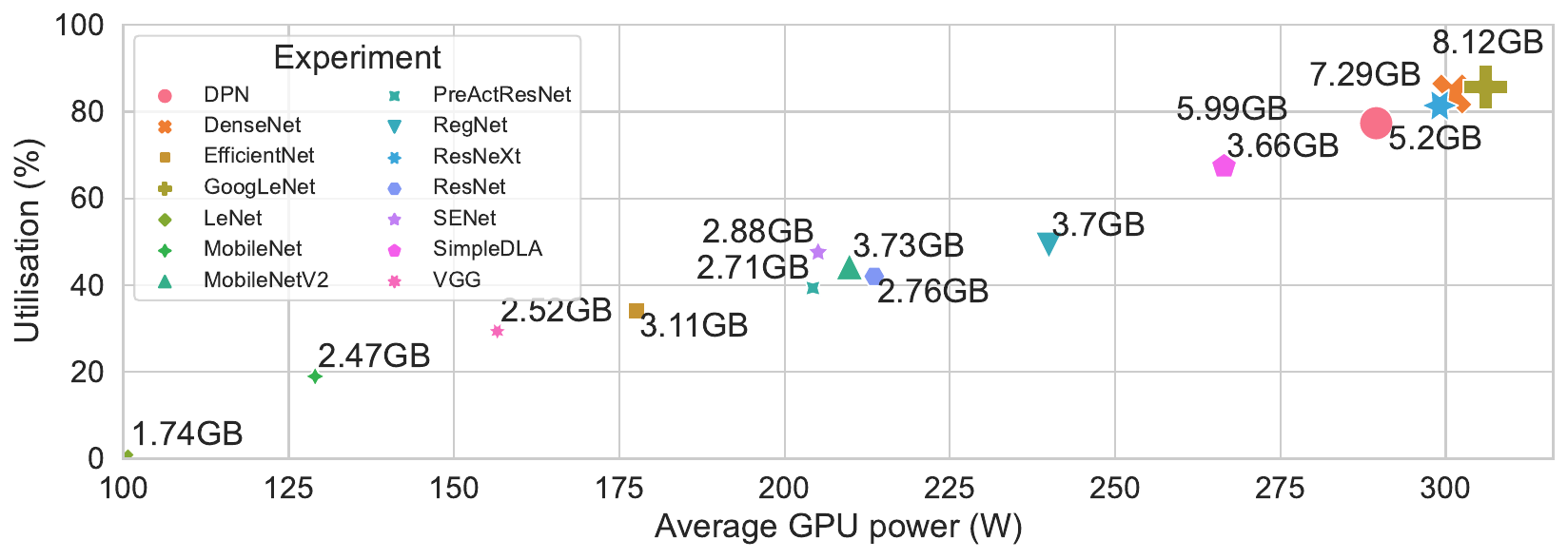}
        \vspace{-6mm}
        \caption{Power usage by model (inference).}
        \label{fig:gpu_ram_testing}
    \end{subfigure}
    \vspace{-3mm}
    \caption{Utilisation and power consumption (considering the GPU RAM usage) - HC-1.}
    \label{fig:gpu_ram}
\end{figure}

Given that CPU and DRAM usage do not change drastically across different models, in Fig.~\ref{fig:gpu_ram}, we present an example of the power consumption as a function of the utilisation and the usage of the GPU's VRAM. For training, a larger GPU VRAM use reflects, most of the time, higher utilisation and increased power consumption. This is even more prominent during inference. Moreover, the results indicate a high correlation between utilisation and power consumption, but up to a certain point (e.g., $\rho \approx 0.81$ for HC-3, $\rho \approx 0.55$ for HC-2). Beyond a power draw of \raisebox{-0.6ex}{\~{}}\SI{300}{\watt}, any further increase did not translate to a dramatic increase in the GPU utilisation. At this point, utilisation was almost $100\%$, so performance was pretty much at its maximum. This is more clear in Fig.~\ref{fig:gpu_ram_training}, where, as said earlier, most models push the GPU to operate close to its TDP. Our results in this investigation are consistent with our previous work~\cite{frost}.

\begin{figure}[t]
    \centering
    \includegraphics[width=0.8\columnwidth]{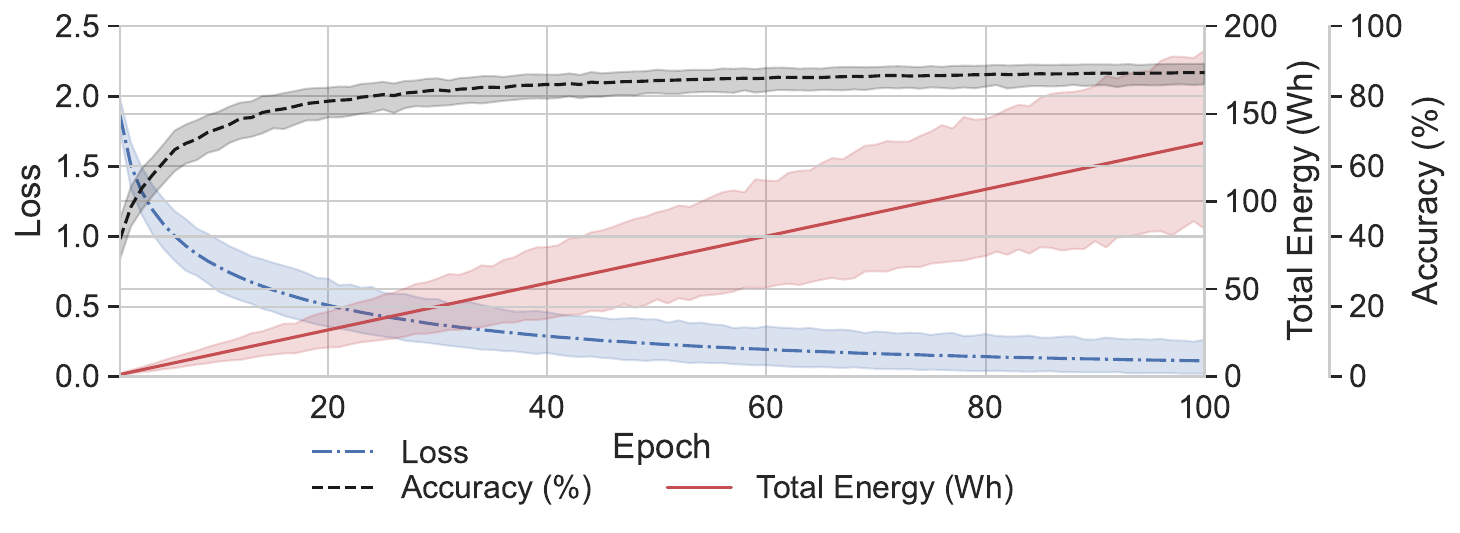}
    \vspace{-3mm}
    \caption{Loss, energy and accuracy per epoch, averaged across all models - the shaded areas show the range of values - HC-3.}
    \label{fig:loss_power}
\end{figure}

We observe a linear relationship by investigating the time and energy consumption, with $r = 0.99$ (e.g., per epoch -- for training, or per X number of samples -- for inference). Our results for that can be found in our repository. However, comparing the model loss, accuracy, and total energy accumulated as the number of epochs increases (average across all models while training -- Fig.~\ref{fig:loss_power}), even though there is no correlation between accuracy and total energy consumed, as the number of epochs increases, the range of values observed for the energy, is greater (relatively) to the accuracy, thus replacing a model can significantly benefit the energy consumption with no significant cost in the accuracy.

\begin{figure}[t]
    \centering
    \begin{subfigure}{0.99\columnwidth}
        \centering
        \includegraphics[width=\linewidth]{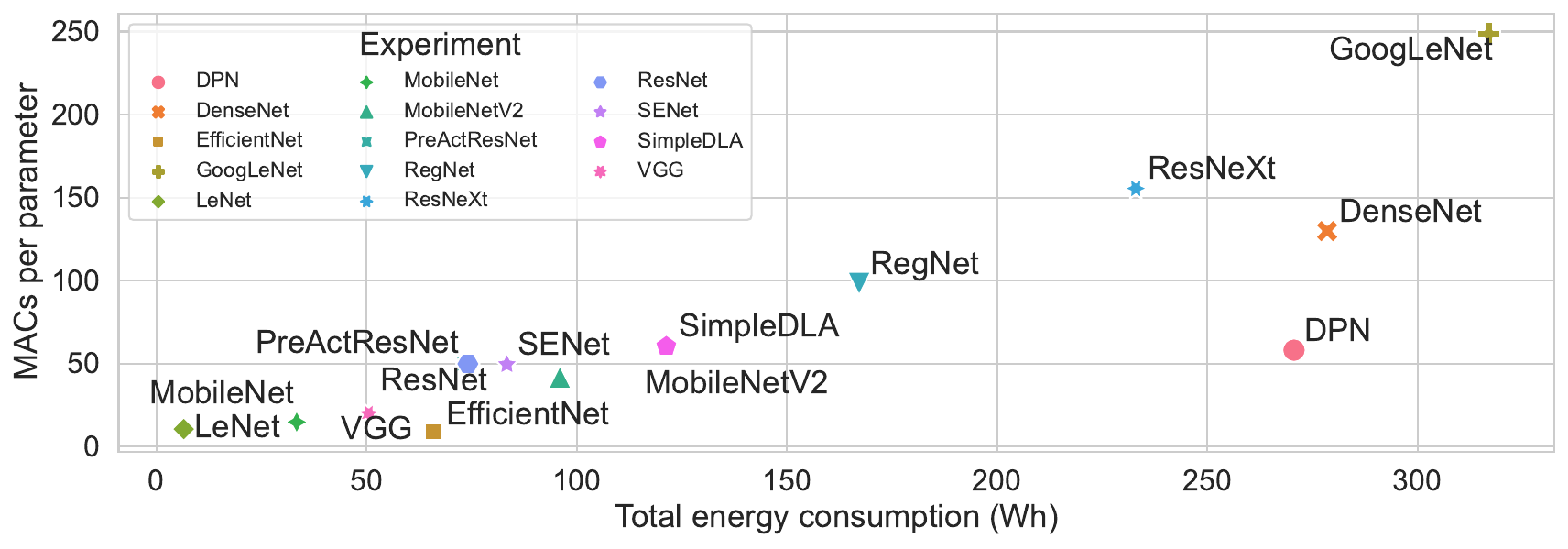}
        \vspace{-6mm}
        \caption{During the training phase.}
        \label{fig:macs_training}
    \end{subfigure}
    \\
    \begin{subfigure}{0.99\columnwidth}
        \centering
        \includegraphics[width=\linewidth]{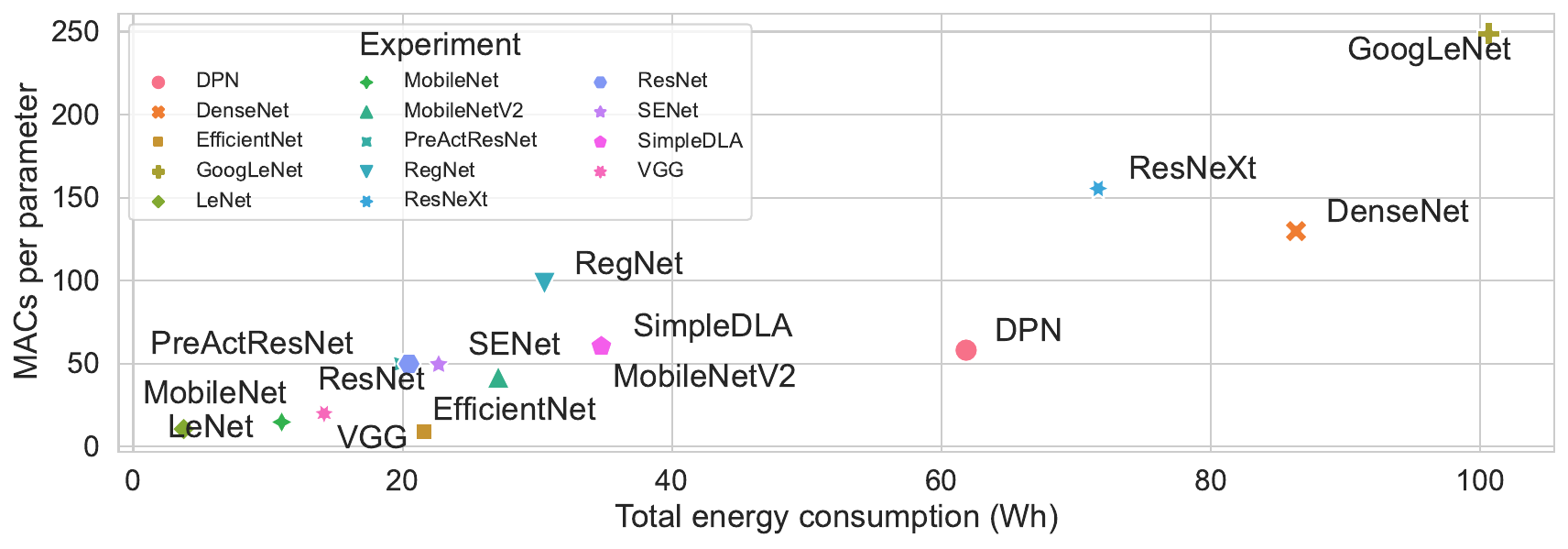}
        \vspace{-6mm}
        \caption{During the inference phase.}
        \label{fig:macs_testing}
    \end{subfigure}
    \vspace{-3mm}
    \caption{Total energy consumption as a function of the MACs per parameter - HC-3.}
    \label{fig:macs_parameters}
\end{figure}

MAC is usually a standard metric that one can use to calculate the complexity of a model and, thus, the expected energy consumption. Comparing the MACs of each model as a function of the total energy, we see a high correlation between them, with $\rho \approx 0.8$ across all HCs (results in our repository). However, our investigation showed that combining MACs with the model (i.e., MACs per model parameter - Fig.~\ref{fig:macs_parameters}) parameters provides a better metric for that. For both training (Fig.~\ref{fig:macs_training}) and inference (Fig.~\ref{fig:macs_testing}), we see a strong correlation across them ($\rho \approx 0.9$ across all HCs).

\begin{figure}[t]
    \centering
    \includegraphics[width=1\columnwidth]{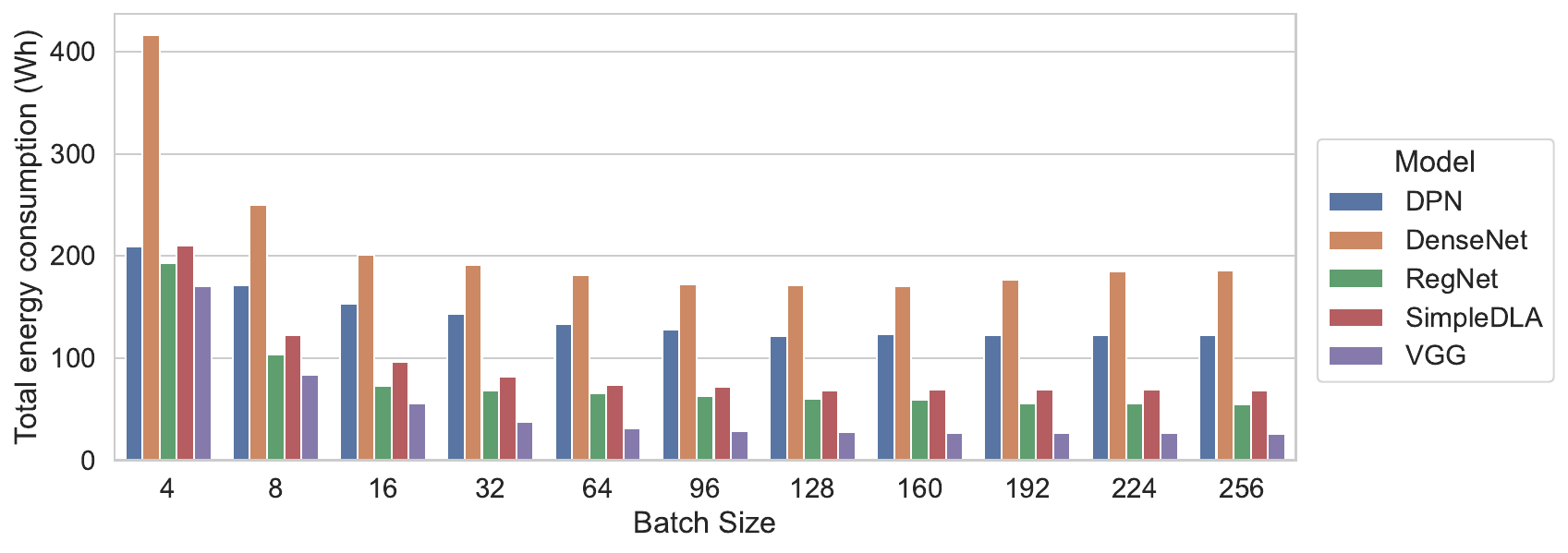}
    \vspace{-8mm}
    \caption{Total energy consumption as a function of the batch size - HC-2.}
    \label{fig:batch}
\end{figure}

Finally, comparing different batch sizes for training and inference (Fig.~\ref{fig:batch}), we see that smaller batch sizes increase the power consumption. There is a direct correlation with the GPU utilisation for each model. For all setups, there is a batch size that minimises the power consumption, with no further improvements shown if the batch size is increased. Considering that smaller batch sizes achieve higher accuracy~\cite{batchSizeComparison}, there is a tradeoff between the accuracy and the energy consumption that can be further investigated.

\section{Discussion}\label{sec:discussion}
Starting with our initial observations (Sec.~\ref{subsec:init_statistics}), it is clear that each model's unique architecture does not allow room for cross-model observations (i.e., if a model's energy consumption is low, there is no obvious way to say that another model will have an equally low energy consumption). Further investigation of the specific architectures and model layers and how they affect energy consumption could identify more similarities. This could be considered a future activity.

Broadly speaking (Fig.~\ref{fig:loss_power}), benefits in energy reduction can outperform the gains in accuracy on many occasions. Moreover, training and inference durations are not correlated; thus, cross-phase or cross-hardware estimations are not promising. Even though a rule of thumb could say that training will require three times more time for the same number of samples, this is not always the case.

Considering that time and total energy are linear, short-living profilers (e.g., training for one epoch or inferring for a small number of samples) can be used to extrapolate the total energy for larger scenarios. In addition, for accuracy and duration, it was evident that models achieving comparable accuracy but ``running faster'' can introduce huge energy benefits in the long term. From what was shown in Fig.~\ref{fig:combined_power} and taking into account Facebook's energy split presented in Sec.~\ref{subsec:energy_consumption}, a less power-hungry model during inference should be prioritised for a pipeline over a less energy-intensive model during training. This observation can be combined with the total energy and time to get even more accurate estimations for both training and inference across different models. Moreover, strategies that analyse the learning curves (from the initial epochs) in conjunction with the power could estimate the expected total energy consumption, identifying models that better fit a given scenario.

As shown in Fig.~\ref{fig:gpu_ram}, hardware devices' power profiles are not exactly linear. Usually, manufacturers push their devices to the limit to squeeze a narrow increase in performance. Smart strategies like the one introduced in~\cite{frost} (power capping optimisations) can exploit that and significantly reduce the total energy consumed. Finally, if an estimation of the model's expected energy is required, contrary to the literature that proposes using the model's MACs, we identified the MACs per model parameter as a more suitable candidate. Similar correlations are observed across all setups, proving it is a uniform solution for accurately estimating the expected energy consumption.

\section{Conclusions}\label{sec:conclusion}
This work presented an extensive analysis across multiple ML models and hardware setups to uncover techniques for improving sustainability without sacrificing effectiveness -- the investigation methodology combined software-based power measurements with tracking of hardware utilisation and model characteristics. The experiments demonstrated that for many models, reductions in energy consumption can outpace marginal accuracy improvements, highlighting the need to balance performance and efficiency. Additionally, assumptions about energy use cannot be reliably made across training and inference or hardware due to a lack of consistent correlations. However, normalising model MACs by the number of parameters provides an excellent indicator of energy consumption in most cases. The insights from this study can guide decisions when constructing ML pipelines, whether choosing architectures and hyperparameters or provisioning hardware resources. There remains ample opportunity for future work to improve sustainability through novel architectures optimised for efficiency and adopting best practices around selective retraining, power capping, and inference-focused model selection. Overall, the evidence clearly shows that with careful planning, ML can continue advancing while aligning with environmental responsibility.

\section*{Acknowledgment}
This work is a contribution by Project REASON, a UK Government funded project under the Future Open Networks Research Challenge (FONRC) sponsored by the Department of Science Innovation and Technology (DSIT). This work was also funded in part by Toshiba Europe Ltd. and Bristol Research and Innovation Laboratory (BRIL).

\bibliography{bib}

\end{document}